\def\year{2017}\relax
\documentclass[letterpaper]{article}
\usepackage{aaai17}
\usepackage{times}
\usepackage{helvet}
\usepackage{courier}
\usepackage{url}
\usepackage{paralist}
\usepackage{amsmath,amsthm}
\usepackage{float}
\usepackage{array}
\usepackage{algorithm}
\usepackage{algorithmic}
\usepackage{graphicx}
\usepackage{multirow}

\DeclareMathOperator{\Rate}{\textsc{Rate}}

\DeclareMathOperator{\NotRate}{\textsc{Not Rate}}
\DeclareMathOperator{\SameUnit}{\textsc{Same Unit}}
\DeclareMathOperator{\NoRelation}{\textsc{No Relation}}
\DeclareMathOperator{\NumUnit}{\textsc{Num Unit}}
\DeclareMathOperator{\DenUnit}{\textsc{Den Unit}}
\DeclareMathOperator{\Vertex}{\textsc{Vertex}}
\DeclareMathOperator{\Edge}{\textsc{Edge}}
\DeclareMathOperator{\Score}{\textsc{Score}}
\DeclareMathOperator{\Label}{\textsc{Label}}
\DeclareMathOperator{\Path}{\textsc{Path}}
\DeclareMathOperator{\pathvar}{\mathrm{path}}

\DeclareMathOperator{\EdgeLabel}{\textsc{EdgeLabel}}
\DeclareMathOperator{\CountMulDiv}{\mathrm{CountMulDiv}}
\DeclareMathOperator{\Decompose}{\textsc{Decompose}}
\DeclareMathOperator{\Joint}{\textsc{Joint}}
\DeclareMathOperator{\Template}{\textsc{Template}}
\DeclareMathOperator{\SingleEq}{\textsc{SingleEq}}
\DeclareMathOperator{\UnitDep}{\textsc{UnitDep}}

\DeclareMathOperator{\Op}{\odot}

\DeclareMathOperator{\LCA}{\textsc{Lca}}
\DeclareMathOperator{\Rel}{\textsc{Irr}}

\DeclareMathOperator{\I}{\mathcal{I}}

\DeclareMathOperator{\Tuples}{\textsc{Tuples}}
\DeclareMathOperator{\Graphs}{\textsc{Graphs}}
\DeclareMathOperator{\Trees}{\textsc{Trees}}

\newcolumntype{L}[1]{>{\raggedright\let\newline\\\arraybackslash\hspace{0pt}}m{#1}}
\newcolumntype{C}[1]{>{\centering\let\newline\\\arraybackslash\hspace{0pt}}m{#1}}
\newcolumntype{R}[1]{>{\raggedleft\let\newline\\\arraybackslash\hspace{0pt}}m{#1}}

\begin{document}
%
\title{Unit Dependency Graph and its Application to Arithmetic Word Problem Solving}
\author{Subhro Roy \and Dan Roth\\
University of Illinois, Urbana Champaign\\
\{sroy9, danr\}@illinois.edu\\
}
\maketitle

\begin{abstract}
Math word problems provide a natural abstraction to a range of natural
language understanding problems that involve reasoning about
quantities, such as interpreting election results, news about
casualties, and the financial section of a newspaper.  Units
associated with the quantities often provide information that is
essential to support this reasoning. This paper proposes a principled
way to capture and reason about units and shows how it can benefit an
arithmetic word problem solver.  This paper presents the concept of
Unit Dependency Graphs (UDGs), which provides a compact representation
of the dependencies between units of numbers mentioned in a given
problem. Inducing the UDG alleviates the brittleness of the unit
extraction system and allows for a natural way to leverage domain
knowledge about unit compatibility, for word problem solving. We
introduce a decomposed model for inducing UDGs with minimal additional
annotations, and use it to augment the expressions used in the
arithmetic word problem solver of (Roy and Roth 2015) via a
constrained inference framework. We show that introduction of UDGs
reduces the error of the solver by over 10\%, surpassing all existing
systems for solving arithmetic word problems. In addition, it also
makes the system more robust to adaptation to new vocabulary and
equation forms .
\end{abstract}

\section{Introduction}
Understanding election results, sport commentaries and financial news,
all require reasoning with respect to quantities. Math word problems
provide a natural abstraction to these quantitative reasoning
problems.  As a result, there has a been a growing interest in
developing methods which automatically solve math word problems
\cite{KHSEA15,KZBA14,RoyRo15,MitraBa16}. 

Units associated with numbers or the question often provide essential
information to support the reasoning required in math word
problems. Consider the arithmetic word problem in Example 1. The units
of ``66'' and ``10'' are both ``flowers'', which indicate they can be
added or subtracted. Although unit of ``8'' is also ``flower'', it is
associated with a rate, indicating the number of flowers in each
bouquet. As a result, ``8'' effectively has unit ``flowers per
bouquet''. Detecting such rate units help understand that ``8'' will
more likely be multiplied or divided to arrive at the
solution. Finally, the question asks for the number of ``bouquets'',
indicating ``8'' will likely be divided, and not multiplied.  Knowing
such interactions could help understand the situation and perform
better quantitative reasoning. In addition, given that unit extraction
is a noisy process, this can make it more robust via global reasoning.

\begin{table}
\centering
\begin{tabular}{|p{7.5cm}|}
\hline
Example 1 \\
\hline
Isabel picked 66 flowers for her friend’s wedding. She was making
bouquets with 8 flowers in each one. If 10 of the flowers wilted
before the wedding, how many bouquets could she still make?\\
\hline
\end{tabular}
\end{table}


In this paper, we introduce the concept of {\em unit dependency graph}
(UDG) for math word problems, to represent the relationships among the
units of different numbers, and the question being asked. We also
introduce a strategy to extract annotations for unit dependency
graphs, with minimal additional annotations. In particular, we use the
answers to math problems, along with the rate annotations for a few
selected problems, to generate complete annotations for unit
dependency graphs. Finally, we develop a decomposed model to predict
UDG given an input math word problem.

We augment the arithmetic word problem solver of \cite{RoyRo15} to
predict a unit dependency graph, along with the solution expression of
the input arithmetic word problem. Forcing the solver to respect the
dependencies of the unit dependency graph enables us to improve unit
extractions, as well as leverage the domain knowledge about unit
dependencies in math reasoning. The introduction of unit dependency
graphs reduced the error of the solver by over $10\%$, while also
making it more robust to reduction in lexical and template overlap of
the dataset.


\section{Unit Dependency Graph}
We first introduce the idea of a generalized rate, and its unit
representation. We define \textbf{rate} to be any quantity which is
some measure corresponding to one unit of some other quantity.  This
includes explicit rates like ``40 miles per hour'', as well as
implicit rates like the one in ``Each student has 3 books''.
Consequently, units for rate quantities take the form {\em ``A per
B''}, where A and B refer to different entities. We refer to A as Num
Unit (short for Numerator Unit), and B as Den Unit (short for
denominator unit). Table \ref{tab:unitsforrates} shows examples of Num
and Den Units for various rate mentions.

\begin{table}
\centering
\small
\begin{tabular}{|p{4cm}|c|c|}
\hline
Mention & Num Unit & Den Unit \\\hline\hline
{\em 40 miles per hour} &  mile & hour \\\hline
{\em Each student has 3 books.} & book & student \\\hline
\end{tabular}
\caption{Units of rate quantities}
\label{tab:unitsforrates}
\end{table} 

\begin{figure*}[ht]
  \centering \small
  \begin{tabular}{|p{5cm}|p{7cm}|p{4cm}|}
    \hline Problem & Unit Dependency Graph & Expression Tree of Solution\\
    \hline
    \vspace{0.2in}
    {\em Isabel picked 66 flowers for her friend’s wedding. She was
      making bouquets with 8 flowers in each one. If 10 of the flowers
      wilted before the wedding, how many bouquets could she still
      make?}
    &
    \begin{center}
      \includegraphics[width=0.3\textwidth]{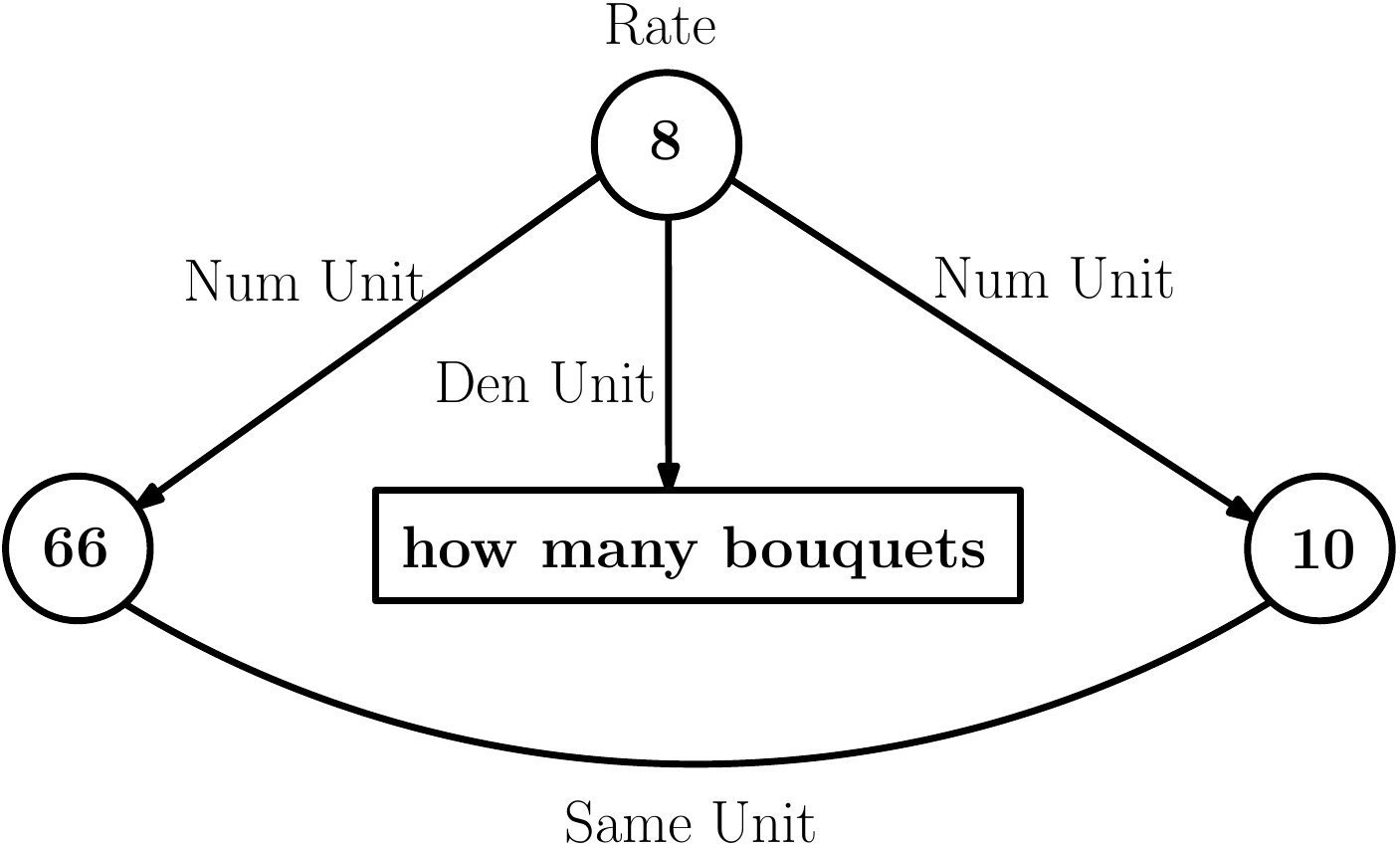}
    \end{center} 
    &
    \vspace{0.2in}
    \begin{center}
      \includegraphics[width=0.15\textwidth]{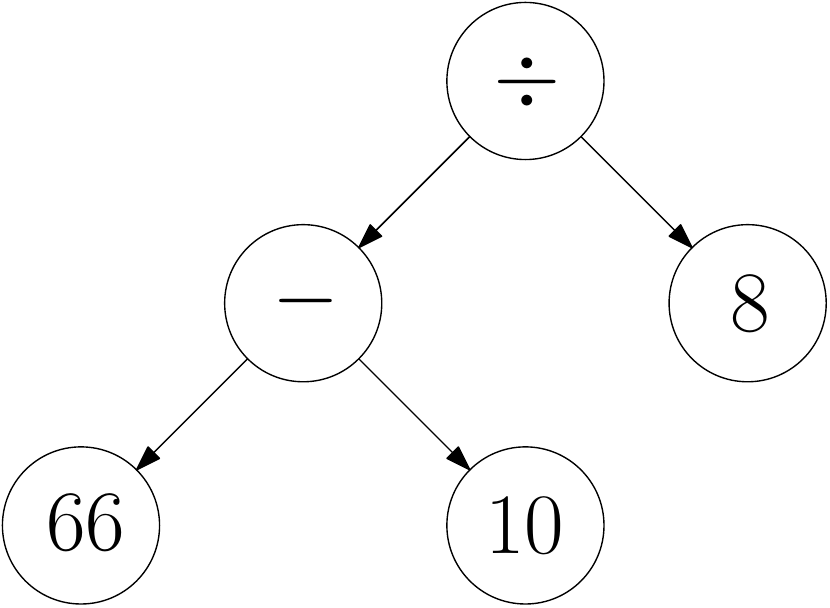}
    \end{center}\\
    \hline
  \end{tabular}
                                   
  \caption{\footnotesize An arithmetic word problem, its UDG, and a
  tree representation of the solution $(66-10)/8$. Several
  dependencies exist between the UDG and the final solution of a
  problem. Here, ``66'' and ``10'' are connected via $\SameUnit$ edge,
  hence they can be added or subtracted, ``8'' is connected by
  $\DenUnit$ to the question, indicating that some expression will be
  divided by ``8'' to get the answer's unit. }

  \label{fig:rug}
\end{figure*}

A unit dependency graph (UDG) of a math word problem is a graph
representing the relations among quantity units and the question
asked. Fig. \ref{fig:rug} shows an example of a math word problem and
its unit dependency graph.  For each quantity mentioned in the
problem, there exists a vertex in the unit dependency graph. In
addition, there is also a vertex representing the question
asked. Therefore, if a math problem mentions $n$ quantities, its unit
dependency graph will have $n+1$ vertices. In the example in
Fig \ref{fig:rug}, there is one vertex corresponding to each of the
quantities $66$, $8$ and $10$, and one vertex representing the
question part ``how many bouquets could she still make ?''.

                                   


A vertex representing a number, is labeled $\Rate$, if the
corresponding quantity describes a rate relationship (according to the
aforementioned definition). In fig \ref{fig:rug}, ``8'' is labeled as
a $\Rate$ since it indicates the number of flowers in each
bouquet. Similarly, a vertex corresponding to the question is marked
$\Rate$ if the question asks for a rate. 

Edges of a UDG can be directed as well as undirected. Each
undirected edge has the label $\SameUnit$, indicating that the
connected vertices have the same unit. Each directed edge going from
vertex $u$ to vertex $v$ can have one of the following labels:

\begin{enumerate}
  \item \textbf{$\NumUnit$} : Valid only for directed edges with source
  vertex $u$ labeled as $\Rate$, indicates that Num Unit of $u$ matches
  the unit of the destination vertex $v$.

  \item \textbf{$\DenUnit$} : Valid only for directed edges with source
  vertex labeled as $\Rate$, indicates that Den Unit of source vertex $u$
  matches the unit of the destination vertex $v$.
\end{enumerate}

If no edge exists between a pair of vertices, they have unrelated units.

Several dependencies exist between the vertex and edge labels of the 
unit dependency graph of a problem, and its solution expression. Sec 
\ref{sec:dependencies} discusses these dependencies and how they
can be leveraged to improve math problem solving.

\section{Learning to Predict UDGs}
Predicting UDG for a math word problem is essentially a structured
prediction problem. However, since we have limited training data, we
develop a decomposed model to predict parts of the structure
independently, and then perform joint inference to enforce coherent
predictions. This has been shown to be an effective method for
structured prediction in the presence of limited
data \cite{PRYZ05,SuttonMc07}. Empirically, we found our decomposed
model to be superior to jointly trained alternatives (see
Section \ref{sec:exp}).

Our decomposed model for UDG prediction uses the following two classifiers.
\begin{enumerate}
\item \textbf{Vertex Classifier} : This is a binary classifier, which
  takes a vertex of the UDG as input, and decides whether it
  denotes a rate.

\item \textbf{Edge Classifier} : This is a multiclass classifier,
  which takes as input a pair of nodes of the UDG, and predicts the
  properties of the edge connecting those nodes.

\end{enumerate}
Finally, a constrained inference module combines the output of the
two classifiers to construct a UDG. We provide details
of the components in the following subsections.

\subsection{Vertex Classifier}

In order to detect rate quantities, we train a binary classifier.
Given problem text $P$ and a vertex $v$ of the UDG, the classifier
predicts whether $v$ represents a rate. It predicts one of two labels
- $\Rate$ or $\NotRate$. The vertex $v$ is either a quantity mentioned
in $P$, or the question of $P$. The features used for the
classification are as follows :
\begin{enumerate}

  \item \textbf{Context Features}: We add unigrams, bigrams, part of
    speech tags, and their conjunctions from the neighborhood of $v$.

  \item \textbf{Rule based Extraction Features}: We add a feature
    indicating whether a rule based approach can detect $v$ as a rate.
\end{enumerate}

\subsection{Edge Classifier}

We train a multiclass classifier to determine the properties of the
edges of the UDG. Given problem text $P$ and a pair of vertices $v_i$
and $v_j$ ($i<j$), the classifier predicts one of the six labels :
\begin{enumerate}
\item \textbf{$\SameUnit$} : Indicates that $v_i$ and $v_j$ should be
  connected by an undirected edge labeled $\SameUnit$.
\item \textbf{$\NoRelation$} : Indicates no edge exists between $v_i$
  and $v_j$.
\item \textbf{$\Rate^{\rightarrow}_{Num}$} : Indicates that
  $v_i$ is a rate, and the Num Unit of $v_i$ matches the unit of
  $v_j$.
\item \textbf{$\Rate^{\leftarrow}_{Num}$} : Indicates that
  $v_j$ is a rate, and the Num Unit of $v_j$ matches the unit of
  $v_i$.
\item We similarly define \textbf{$\Rate^{\rightarrow}_{Den}$}
  and \textbf{$\Rate^{\leftarrow}_{Den}$}.
\end{enumerate}
\noindent
The features used for the classification are :
\begin{enumerate}
  \item \textbf{Context Features}: For each vertex $v$ in the query,
    we add the context features described for Vertex classifier.

  \item \textbf{Rule based Extraction Features}: We add a feature
    indicating whether each of the queried vertices is detected
    as a rate by the rule based system. In addition, we also add 
    features denoting whether there are common tokens in the units 
    of $v_i$ and $v_j$. 

\end{enumerate}

\subsection{Constrained Inference}

Our constrained inference module takes the scores of the Vertex and
Edge classifiers, and combines them to find the most probable unit
dependency graph for a problem. We define $\Vertex(v, l)$ to be the score
predicted by the Vertex classifier for labeling vertex $v$ of a UDG
with label $l$, where $l \in \{\Rate, \NotRate\}$. Similarly, we
define $\Edge(v_i, v_j, l)$ to be the score predicted by the Edge
classifier for the assignment of label $l$ to the edge between $v_i$
and $v_j$. Here the label $l$ is one of the six labels defined for the
edge classifier.

Let $G$ be a UDG with vertex set $V$. We define the score for $G$ as
follows:
\begin{align*}
&\Score(G) = \sum_{\substack{v \in V \\ \Label(G, v) = \Rate}} \Vertex(v, \Rate) + \\
& \lambda \times \sum_{v_i, v_j \in V, i<j} \Edge(v_i, v_j, \Label(G, v_i, v_j))
\end{align*}
\noindent
where $\lambda$ is a scaling factor, and $\Label$ maps labels of the
UDG, to the labels of the corresponding
classifiers. $\Label(G,v)$ maps to $\Rate$, if $v$ is a rate,
otherwise it maps to $\NotRate$. Similarly, if no edge exists between
$v_i$ and $v_j$, $\Label(G, v_i, v_j)$ maps to $\NoRelation$, if Num
Unit of $v_i$ matches the unit of $v_j$, $\Label(G, v_i, v_j)$ maps to
$\Rate^{\rightarrow}_{Num}$, and so on. Finally, the inference problem
has the following form: 
\begin{align*}
\arg\max_{G \in \Graphs} \Score(G)
\end{align*}
where $\Graphs$ is the set of all valid unit dependency graphs for the
input problem.

\section{Joint Inference With An Arithmetic Solver}
In this section, we describe our joint inference procedure to predict
both a UDG and the solution of an input arithmetic word problem. Our
model is built on the arithmetic word problem solver
of \cite{RoyRo15}, and we briefly describe it in the following
sections. We first describe the concept of expression trees, and next
describe the solver, which leverages expression tree representation of
the solutions.

\subsection{Monotonic Expression Tree}
\label{sec:monotonic}

An \textbf{expression tree} is a binary tree representation of a
mathematical expression, where leaves represent numbers, and all
non-leaf nodes represent operations. Fig \ref{fig:rug} shows an
example of an arithmetic word problem and the expression tree of the
solution mathematical expression.  A \textbf{monotonic expression
tree} is a normalized expression tree representation for math
expressions, which restricts the order of combination of addition and
subtraction nodes, and multiplication and division nodes. The
expression tree in Fig \ref{fig:rug} is monotonic.






\subsection{Arithmetic Word Problem Solver}
\label{sec:arithmetic}

We now describe the solver pipeline of \cite{RoyRo15}. Given a problem
$P$ with quantities $q_1, q_2, \ldots, q_n$, the solver uses the
following two classifiers.

\begin{enumerate}

\item \textbf{Irrelevance Classifier} : Given as input, problem $P$
and quantity $q_i$ mentioned in $P$, the classifier decides whether
$q_i$ is irrelevant for the solution. The score of this classifier is
denoted as $\Rel(q)$.

\item \textbf{LCA Operation Classsifier} : Given as input, problem
$P$ and a pair of quantities $q_i$ and $q_j$ $(i < j)$, the classifier
predicts the operation at the lowest common ancestor (LCA) node of
$q_i$ and $q_j$, in the solution expression tree of problem $P$. The
set of possible operations are $+$, $-$, $-_r$, $\times$, $\div$ and
$\div_r$ (the subscript $r$ indicates reverse order).  Considering
only monotonic expression trees for the solution makes this operation
unique for any pair of quantities. The score of this classifier for
operation $o$ is denoted as $\LCA(q_i, q_j, o)$.

\end{enumerate}

The above classifiers are used to gather irrelevance scores for each 
number, and LCA operation scores for each pair of numbers. Finally,
constrained inference procedure combines these scores to generate the
solution expression tree.

Let $\I(T)$ be the set of all quantities in $P$ which are not used in
expression tree $T$, and $\lambda_{\Rel}$ be a scaling parameter. The
score $\Score(T)$ of an expression tree $T$ is defined as:
\begin{align*}
  & \Score(T) = 
  \lambda_{\Rel} \sum_{q \in \I(T)} \Rel(q) +    \\
  & \sum_{q_i,q_j \notin \I(T)} \LCA(q_i, q_j, \Op_{LCA}(q_i, q_j, T))
\end{align*}
\noindent
where $\Op_{LCA}(q_i, q_j, T)$ denotes the operation at the lowest
common ancestor node of $q_i$ and $q_j$ in monotonic expression tree
$T$. Let $\Trees$ be the set of valid expressions that can be formed
using the quantities in a problem $P$, and also give positive
solutions. The inference algorithm now becomes:
\begin{align*} 
\arg\max_{T \in \Trees} \Score(T) 
\end{align*}

\subsection{Joint Inference}

We combine the scoring functions of UDG prediction and the
ones from the solver of \cite{RoyRo15}, so that we can jointly predict
the UDG and the solution of the problem. For an input
arithmetic word problem $P$, we score tuples $(G, T)$ (where $G$ is a
candidate UDG for $P$, and $T$ is a candidate solution 
expression tree of $P$) as follows :
\begin{align*}
  & \Score(G, T) = \lambda_{\Rel} \sum_{q \in \I(T)} \Rel(q) +    \\
  & \sum_{q_i,q_j \notin \I(T)} \LCA(q_i, q_j, \Op_{LCA}(q_i, q_j, T)) + \\
  & \lambda_{\Vertex} \sum_{\substack{v \in V \\ \Label(G, v) = \Rate}} \Vertex(v, \Rate) + \\
  & \lambda_{\Edge} \sum_{v_i, v_j \in V, i<j} \Edge(v_i, v_j, \Label(G, v_i, v_j))
\end{align*}
\noindent
where $\lambda_{\Rel}$, $\lambda_{\Vertex}$ and $\lambda_{\Edge}$ are
scaling parameters. This is simply a scaled addition of the scores for
UDG prediction and solution expression
generation. Finally, the inference problem is
\begin{align*}
\arg\max_{(G, T) \in \Tuples} \Score(G, T)
\end{align*}
\noindent
where $\Tuples$ is the set of all tuples $(G, T)$, such that
$G \in \Graphs$, $T \in \Trees$, and $G$ is a consistent UDG for the
solution tree $T$.

\subsection{Consistent Rate Unit Graphs}
\label{sec:dependencies}

We have a set of conditions to check whether $G$ is a consistent
UDG for monotonic tree $T$. Most of these conditions are
expressed in terms of $\Path(T,v_i, v_j)$, which takes as input a pair
of vertices $v_i, v_j$ of the UDG $G$, and a monotonic
expression tree $T$, and returns the following.
\begin{enumerate}

\item If both $v_i$ and $v_j$ are numbers, and their corresponding
leaf nodes in $T$ are $n_i$ and $n_j$ respectively, then it returns the
nodes in the path connecting $n_i$ and $n_j$ in $T$.

\item If only $v_i$ denotes a number (implying $v_j$ represents the question), 
the function returns the nodes in the path from $n_i$ to the root of
$T$, where $n_i$ is the corresponding leaf node for $v_i$.
 
\end{enumerate}
For the unit dependency graph and solution tree $T$ of Fig \ref{fig:rug},
$\Path(T, 66, 8)$ is $\{-, \div\}$, whereas $\Path(T,
8, \text{question})$ is $\{\div\}$.  Finally, the conditions for
consistency between a UDG $G$ and an expression tree $T$
are as follows:
\begin{enumerate}
\item If $v_i$ is the only vertex labeled $\Rate$ and it is the question,
there should not exist a path from some leaf $n$ to the root of $T$
which has only addition, subtraction nodes. If that exists, it implies
$n$ can be added or subtracted to get the answer, that is, the
corresponding vertex for $n$ in $G$ has same unit as the question, and
should have been labeled $\Rate$.

\item If $v_i$ is labeled $\Rate$ and the question is not,
the path from $n_i$ (corresponding leaf node for $v_i$) to the root of
$T$ cannot have only addition, subtraction nodes. Otherwise, the
question will have same rate units as $v_i$.

\item We also check whether the edge labels are consistent with the vertex
labels using Algorithm \ref{algo:edgelabel}, which computes edge
labels of UDGs, given the expression tree $T$, and vertex
labels.  It uses heuristics like if a rate $r$ is being multiplied by
a non-rate number $n$, the Den Unit of $r$ should match the unit of
$n$, etc. 

\end{enumerate} 
\begin{algorithm}[ht]
  \caption{$\EdgeLabel$}
  \label{algo:edgelabel}
  \small
  \begin{algorithmic}[1]
  \REQUIRE {Monotonic expression tree $T$, vertex pairs $v_i, v_j$, and
  their corresponding vertex labels}
  \ENSURE Label of edge between $v_i$ and $v_j$
  \STATE $\pathvar \leftarrow \Path(T, v_i, v_j)$ 
  \STATE $\CountMulDiv \leftarrow$ Number of Multiplication and Division nodes
  in $\pathvar$   
  \IF{$v_i$ and $v_j$ have same vertex label, and $\CountMulDiv = 0$}
    \RETURN $\SameUnit$
  \ENDIF
  \IF{$v_i$ and $v_j$ have different vertex labels, and $\CountMulDiv = 1$}
    \IF{$\pathvar$ contains $\times$ and $v_i$ is $\Rate$}
      \RETURN $\Rate^{\rightarrow}_{Den}$
    \ENDIF
    \IF{$\pathvar$ contains $\times$ and $v_j$ is $\Rate$}
      \RETURN $\Rate^{\leftarrow}_{Den}$
    \ENDIF
    \IF{$\pathvar$ contains $\div$ and $v_i$ is $\Rate$}
      \RETURN $\Rate^{\rightarrow}_{Num}$
    \ENDIF
    \IF{$\pathvar$ contains $\div_r$ and $v_j$ is $\Rate$}
      \RETURN $\Rate^{\leftarrow}_{Num}$
    \ENDIF
  \ENDIF
  \RETURN Cannot determine edge label
\end{algorithmic}
\end{algorithm}

These consistency conditions prevent the inference procedure from
considering any inconsistent tuples. They help the solver to get rid
of erroneous solutions which involve operations inconsistent with all
high scoring UDGs.

Finally, in order to find the highest scoring consistent tuple, we
have to enumerate the members of $\Tuples$, and score them. The size
of $\Tuples$ however is exponential in the number of quantities in the
problem. As a result, we perform beam search to get the highest
scoring tuple. We first enumerate the members of $\Trees$, and next
for each member of $\Trees$, we enumerate consistent UDGs.

\section{Experiments}
\label{sec:exp}

\subsection{Dataset}

Existing evaluation of arithmetic word problem solvers has several
drawbacks. The evaluation of \cite{RoyRo15} was done separately on
different types of arithmetic problems. This does not capture how well
the systems can distinguish between these different problem
types. Datasets released by \cite{RoyRo15} and \cite{KHSEA15}
mention irrelevant quantities in words, and only the relevant
quantities are mentioned in digits. This removes the challenge of
detecting extraneous quantities.

In order to address the aforementioned issues, we pooled arithmetic
word problems from all available datasets
\cite{HHEK14,RoyRo15,KHSEA15}, and normalized all mentions of
quantities to digits. We next prune problems such that there do not
exist a problem pair with over $80 \%$ match of unigrams and
bigrams. The threshold of $80 \%$ was decided manually by determining
that problems with around $80 \%$ overlap are sufficiently
different. We finally ended up with $831$ problems. We refer to this
dataset as \textbf{AllArith}.

We also create subsets of \textbf{AllArith} using the MAWPS system
\cite{KRAKH16}. MAWPS can generate subsets of word problems based on
lexical and template overlap. Lexical overlap is a measure of reuse of
lexemes among problems in a dataset. High lexeme reuse allows for
spurious associations between the problem text and a correct solution
\cite{KHSEA15}. Evaluating on low lexical overlap subset of the
dataset can show the robustness of solvers to lack of spurious
associations. Template overlap is a measure of reuse of similar
equation templates across the dataset. Several systems focus on
solving problems under the assumption that similar equation templates
have been seen at training time. Evaluating on low template overlap
subset can show the reliance of systems on the reuse of equation
templates. We create two subsets of $415$ problems each - one with low
lexical overlap called \textbf{AllArithLex}, and one with low template
overlap called \textbf{AllArithTmpl}.

We report random $5$-fold cross validation results on all these datasets.
For each fold, we choose $20\%$ of the training data as development set,
and tune the scaling parameters on this set. Once the parameters are set,
we retrain all the models on the entire training data. We use a beam size
of $200$ in all our experiments.

\subsection{Data Acquisition}

In order to learn the classifiers for predicting vertex and edge
labels for UDGs, we need annotated data. However,
gathering vertex and edge labels for UDGs of problems, can
be expensive. In this section, we show that vertex labels for a subset
of problems, along with annotations for solution expressions, can be
sufficient to gather high quality annotations for vertex and edge
labels of UDGs. 

Given an arithmetic word problem $P$, annotated with the monotonic
expression tree $T$ of the solution expression, we try to acquire
annotations for the UDG of $P$. First, we try to determine the labels
for the vertices, and next the edges of the graph.

We check if $T$ has any multiplication or division node. If no such
node is present, we know that all the numbers in the leaves of $T$
have been combined via addition or subtraction, and hence, none of
them describes a rate in terms of the units of other numbers. This
determines that none of $T$'s leaves is a rate, and also, the question
does not ask for a rate. If a multiplication or division node is
present in $T$, we gather annotations for the numbers in the leaves of
$T$ as well as the question of $P$. Annotators were asked to mark
whether each number represents a rate relationship, and whether the
question in $P$ asks for a rate. This process determines the labels
for the vertices of the UDG. Two annotators performed these
annotations, with an agreement of 0.94(kappa).

Once we have the labels for the vertices of the UDG, we
try to infer the labels for the edges using Algorithm
\ref{algo:edgelabel}. When the algorithm is unable to infer the label
for a particular edge, we heuristically label that edge to be
$\NoRelation$.

The above process allowed us to extract high quality annotations for
UDGs with minimal manual annotations. In particular, we
only had to annotate vertex labels for $300$ problems, out of the
$831$ problems in \textbf{AllArith}. Obviously some of the extracted
$\NoRelation$ edge labels are noisy; this can be remedied by
collecting annotations for these cases. However, in this work, we did
not use any manual annotations for edge labels.


\begin{table*}
  \centering
  \footnotesize
  \begin{tabular}{|l||c|c|c||c|c|c||}
    \hline
    \multirow{2}{*}{Features} & \multicolumn{3}{c||}{Vertex Classifier} 
    & \multicolumn{3}{c||}{Edge Classifier} \\\cline{2-7} 
    & AllArith & AllArithLex & AllArithTmpl & AllArith & AllArithLex 
    & AllArithTmpl \\\hline\hline
    All features & 96.7 & 96.2 & 97.5 & 87.1 & 84.3 & 86.6 \\\hline 
    No rule based features & 93.2 & 92.5 & 92.6 & 79.3 & 75.4 & 78.0 \\\hline
    No context features & 95.1 & 94.1 & 95.3 & 78.6 & 70.3 & 75.5 \\\hline
  \end{tabular}
  \caption{\footnotesize  Performance of system components for predicting
  vertex and edge labels for unit dependency graphs}
  \label{tab:modules16}
\end{table*}

\subsection{UDG Prediction}

Table \ref{tab:modules16} shows the performance of the classifiers and
the contribution of each feature type. The results indicate that
rule-based techniques are not sufficient for robust extraction, there
is a need to take context into account. Table \ref{tab:udg} shows the
performance of our decomposed model ($\Decompose$) in correctly
predicting UDGs, as well as the contribution of constraints in the
inference procedure.  Having explicit constraints for the graph
structure provides 3-5\% improvement in correct UDG prediction. 

We also compare against a jointly trained model ($\Joint$), which
learns to predict all vertex and edge labels together. Note that
$\Joint$ also uses the same set of constraints as $\Decompose$ in the
inference procedure, to ensure it only predicts valid unit dependency
graphs.  We found that $\Joint$ does not outperform $\Decompose$,
while taking significantly more time to train. The worse performance
of joint learning is due to: (1) search space being too large for the
joint model to do well given our relatively small dataset size, and
(2) our independent classifiers being good enough, thus supporting
better joint inference. This tradeoff is strongly supported in the
literature \cite{PRYZ05,SuttonMc07}.

Note, that all these evaluations are based on noisy edges
annotations. This was done to reduce further annotation effort.  Also,
less than 15\% of labels were noisy (indicated by fraction of
$\NoRelation$ labels), which makes this evaluation reasonable.

\begin{table}
  \centering
  \footnotesize
  \begin{tabular}{|l|c|c|c|}
    \hline
    & AllArith & AllArithLex & AllArithTmpl \\\hline\hline
    $\Decompose$ & 73.6 & 67.7 & 68.7 \\
    \quad\quad -  constraints &  70.9 &  62.9 & 65.5 \\\hline
    $\Joint$ & 72.9 & 66.7 & 68.4\\\hline 
  \end{tabular}
  \caption{\footnotesize  Performance in predicting UDGs}
  \label{tab:udg}
\end{table}

\subsection{Solving Arithmetic Word Problems}

Here we evaluate the accuracy of our system in correctly solving
arithmetic word problems. We refer to our system as $\UnitDep$.  We
compare against the following systems:

\begin{enumerate}
\item LCA++ : System of \cite{RoyRo15} with feature set augmented
  by neighborhood features, and with only positive answer constraint.
  We found that augmenting the released feature set with context
  features, and removing the integral answer constraint, were helpful.
  Our system $\UnitDep$ also uses the augmented feature set for
  Relevance and LCA operation classifiers, and only positive
  constraint for final solution value.
\item $\Template$ : Template based algebra word problem solver of \cite{KZBA14}.
\item $\SingleEq$ : Single equation word problem solver of \cite{KHSEA15}. 
\end{enumerate}

In order to quantify the gains due to vertex and edge information of
UDGs, we also run two variants of $\UnitDep$ - one with
$\lambda_{\Vertex} = 0$, and one with $\lambda_{\Edge} = 0$.  Table
\ref{tab:solving} shows the performance of these systems on AllArith,
AllArithLex and AllArithTmpl. 
\begin{table}
  \centering
  \footnotesize
  \begin{tabular}{|l||c|c|c||}
    \hline
    System & AllArith & AllArithLex & AllArithTmpl \\\hline\hline
    $\Template$ & 73.7 & 65.5 & 71.3 \\\hline
    $\SingleEq$ & 60.4 & 51.5 & 51.0 \\\hline
    LCA++ & 79.4 & 63.6 & 74.7 \\\hline\hline
    $\UnitDep$ & \textbf{81.7} & \textbf{68.9} & \textbf{79.5} \\\hline
    \quad$\lambda_{\Vertex} = 0$ & 80.3 & 67.2 & 77.1 \\ \hline
    \quad$\lambda_{\Edge} = 0$ & 79.9 & 64.1 & 75.7\\ \hline 
  \end{tabular}
  \caption{\footnotesize  Performance in 
    solving arithmetic word problems}
  \label{tab:solving}
\end{table}

$\UnitDep$ outperforms all other systems across all datasets. Setting
either $\lambda_{\Vertex} = 0$ or $\lambda_{\Edge} = 0$ leads to a
drop in performance, indicating that both vertex and edge information
of UDGs assist in math problem solving. Note that setting both
$\lambda_{\Vertex}$ and $\lambda_{\Edge}$ to $0$, is equivalent to
LCA++. $\SingleEq$ performs worse than other systems, since it does
not handle irrelevant quantities in a problem.

In general, reduction of lexical overlap adversely affects the
performance of most systems. The reduction of template overlap does
not affect performance as much. This is due to the limited number of
equation templates found in arithmetic problems. Introduction of UDGs
make the system more robust to reduction of both lexical and template
overlap. In particular, they provide an absolute improvement of $5\%$
in both AllArithLex and allArithTmpl datasets (indicated by difference
of LCA++ and $\UnitDep$ results).

For the sake of completeness, we also ran our system on the previously
used datasets, achieving $1$\% and $4$\% absolute improvements over
LCA++, in the Illinois dataset \cite{RoyViRo15} and the Commoncore
dataset \cite{RoyRo15} respectively.

\subsection{Discussion}
Most of gains of $\UnitDep$ over LCA++ came from problems where LCA++
was predicting an operation or an expression that was inconsistent
with the units.  A small gain ($10\%$) also comes from problems where
UDGs help detect certain irrelevant quantities, which LCA++ cannot
recognize. Table \ref{tab:correct} lists some of the examples which
$\UnitDep$ gets correct but LCA++ does not.

Most of the mistakes of $\UnitDep$ were due to extraneous quantity
detection (around $50\%$). This was followed by errors due to the lack
of math understanding (around $23\%$). This includes comparison
questions like ``How many more pennies does John have?''. 

\begin{table*}
\centering
\footnotesize
\begin{tabular}{|p{10cm}|c|c|}
\hline 
Problem & LCA++ & $\UnitDep$ \\ \hline\hline  
At lunch a waiter had 10 customers and 5 of them didn't leave a
tip. If he got \$3.0 each from the ones who did tip, how much money
did he earn?  &
10.0-(5.0/3.0) &  3.0*(10.0-5.0)\\\hline 

The schools debate team had 26 boys and 46 girls on it. If they
were split into groups of 9, how many groups could they make? &
9*(26+46) &  (26+46)/9 \\\hline

Melanie picked 7 plums and 4 oranges from the orchard . She gave
3 plums to Sam . How many plums does she have now ? &
 (7+4)-3 & (7-3)\\\hline

Isabella’s hair is 18.0 inches long. By the end of the year her hair
is 24.0 inches long. How much hair did she grow? &
(18.0*24.0) & (24.0-18.0)\\\hline

\end{tabular}
\caption{Examples of problems which $\UnitDep$ gets correct, but LCA++ does not.}
\label{tab:correct}
\end{table*}

\section{Related Work}
There has been a recent interest in automatically solving math word
problems. \cite{HHEK14,MitraBa16} focus on addition-subtraction problems,
\cite{RoyViRo15} look at single operation problems, \cite{RoyRo15}
as well as our work look at arithmetic problems with each number in
question used at most once in the answer, \cite{KHSEA15} focus on
single equation problems, and finally \cite{KZBA14} focus on algebra
word problems. None of them explicitly model the relations between
rates, units and the question asked. In contrast, we model these
relations via unit dependency graphs.  Learning to predict these
graphs enables us to gain robustness over rule-based extractions.
Other than those related to math word problems, there has been some
work in extracting units and rates of
quantities \cite{RoyViRo15,Kuehne04a,Kuehne04b}. All of them employ
rule based systems to extract units, rates and their relations.

\section{Conclusion}
In this paper, we introduced the concept of unit dependency graphs, to
model the dependencies among units of numbers mentioned in a math word
problem, and the question asked. The dependencies of UDGs help improve
performance of an existing arithmetic word problem solver, while also
making it more robust to low lexical and template overlap of the
dataset. We believe a similar strategy can be used to incorporate
various kinds of domain knowledge in math word problem solving. Our
future directions will revolve around this, particularly to
incorporate knowledge of entities, transfers and math concepts. Code
and dataset are available at {\em
http://cogcomp.cs.illinois.edu/page/publication\_view/804}.

\section*{Acknowledgements}
This work is funded by DARPA under agreement number FA8750-13-2-0008,
and a grant from the Allen Institute for Artificial Intelligence
(allenai.org).

\bibliography{ccg,cited,subhro}
\bibliographystyle{aaai}

\end{document}